%% file: main.tex
\documentclass[10pt,twocolumn,letterpaper]{article}

\usepackage{color}
\usepackage{colortbl}
\usepackage[dvipsnames]{xcolor}         % colors
\usepackage{adjustbox}
\usepackage{soul}
\definecolor{mygray}{gray}{.9}
\definecolor{Gray}{gray}{0.8}
\definecolor{LGray}{gray}{0.9}
\sethlcolor{LGray}
\usepackage{graphicx}
\usepackage{caption}
\usepackage{subcaption}

\usepackage[camera]{iccv}      % To produce the REVIEW version
\input{preamble}

\definecolor{iccvblue}{rgb}{0.21,0.49,0.74}
\usepackage[pagebackref,breaklinks,colorlinks,allcolors=iccvblue]
{hyperref}

\def\paperID{12} % *** Enter the Paper ID here
\def\confName{ICCV}
\def\confYear{2025}

\title{
  \adjustbox{valign=c}{\includegraphics[height=24pt]{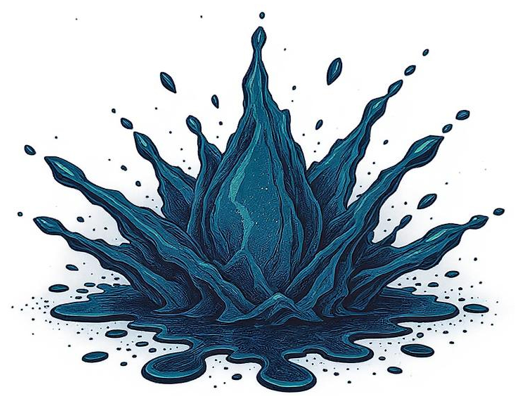}}~%
  SynSpill: Improved Industrial Spill Detection With Synthetic Data
}
\author{Aaditya Baranwal\textsuperscript{$\dagger$}\\
University of Central Florida\\
\small Orlando, FL, USA\\
{\tt\small aaditya.baranwal@ucf.edu}
\and
Abdul Mueez\\
University of Central Florida\\
\small Orlando, FL, USA\\
{\tt\small abdul.mueez@iitj.ac.in}
\and
Jason Voelker\\
Siemens Energy\\
\small Orlando, FL, USA\\
{\tt\small jason.voelker@siemens-energy.com}
\and
Guneet Bhatia\\
Siemens Energy\\
\small Orlando, FL, USA\\
{\tt\small guneet.bhatia@siemens-energy.com}
\and
Shruti Vyas\\
University of Central Florida\\
\small Orlando, FL, USA\\
{\tt\small shruti@ucf.edu}
}

\begin{document}

\twocolumn[{
\maketitle
\vspace{-1em}
\begin{center}
    \includegraphics[width=0.98\textwidth]{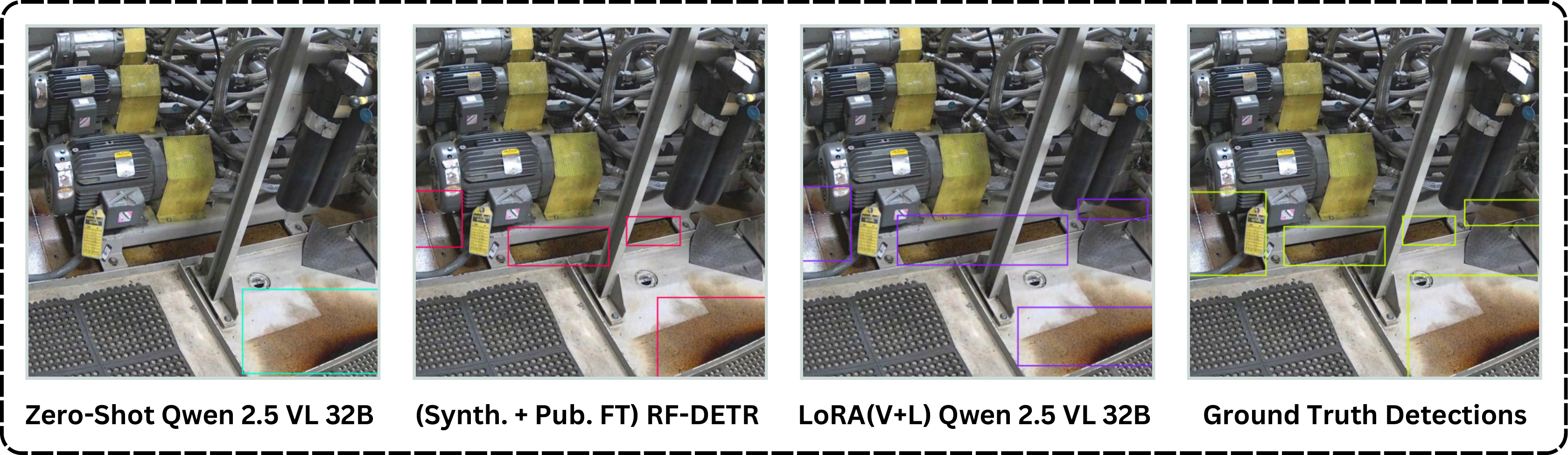}
    \captionof{figure}{\textbf{Comparative detection performance of competing methods on a real-world CCTV image of an industrial spill.} 
    We visualize and contrast the predictions of three models: (1) a Zero-Shot Qwen2.5-VL-32B baseline without adaptation, 
    (2) a PEFT-adapted Qwen2.5-VL-32B using LoRA on synthetic and web-scraped public data, 
    and (3) a finetuned RF-DETR Base model trained on the same hybrid dataset. 
    The ground-truth annotation is overlaid for reference. 
    This qualitative comparison highlights the relative gains.
    }
    \label{fig:graphical-abstract}
\end{center}
\vspace{1em}
}]

\input{sec/0_abstract}
\input{sec/1_intro}
\input{sec/2_relatedwork}

\input{sec/3_setup}

\input{sec/4_synthetic}

\input{sec/5_experiments}

\input{sec/6_conclusion}
\input{sec/7_limitations}
\newpage
{
    \small
    \bibliographystyle{ieeenat_fullname}
    \bibliography{main}
}

\input{sec/X_suppl}

\end{document}

%% file: preamble.tex
\usepackage{xcolor}

%% file: sec/0_abstract.tex
\begin{abstract}
Large-scale Vision-Language Models (VLMs) have transformed general-purpose visual recognition through strong zero-shot capabilities. However, their performance degrades significantly in niche, safety-critical domains such as industrial spill detection, where hazardous events are rare, sensitive, and difficult to annotate. This scarcity---driven by privacy concerns, data sensitivity, and the infrequency of real incidents---renders conventional fine-tuning of detectors infeasible for most industrial settings.

We address this challenge by introducing a scalable framework centered on a high-quality synthetic data generation pipeline. We demonstrate that this synthetic corpus enables effective \textit{Parameter-Efficient Fine-Tuning (PEFT)} of VLMs and substantially boosts the performance of state-of-the-art object detectors such as YOLO and DETR. Notably, in the absence of synthetic data (SynSpill dataset), VLMs still generalize better to unseen spill scenarios than these detectors. When SynSpill is used, both VLMs and detectors achieve marked improvements, with their performance becoming comparable.

Our results underscore that high-fidelity synthetic data is a powerful means to bridge the domain gap in safety-critical applications. The combination of synthetic generation and lightweight adaptation offers a cost-effective, scalable pathway for deploying vision systems in industrial environments where real data is scarce/impractical to obtain.

Project Page: \href{https://synspill.vercel.app}{synspill.vercel.app}
\end{abstract}

%% file: sec/1_intro.tex
\vspace{-2em}
\section{Introduction}
\label{sec:intro}

\begin{figure}[t]
    \centering
        \includegraphics[width=\linewidth]{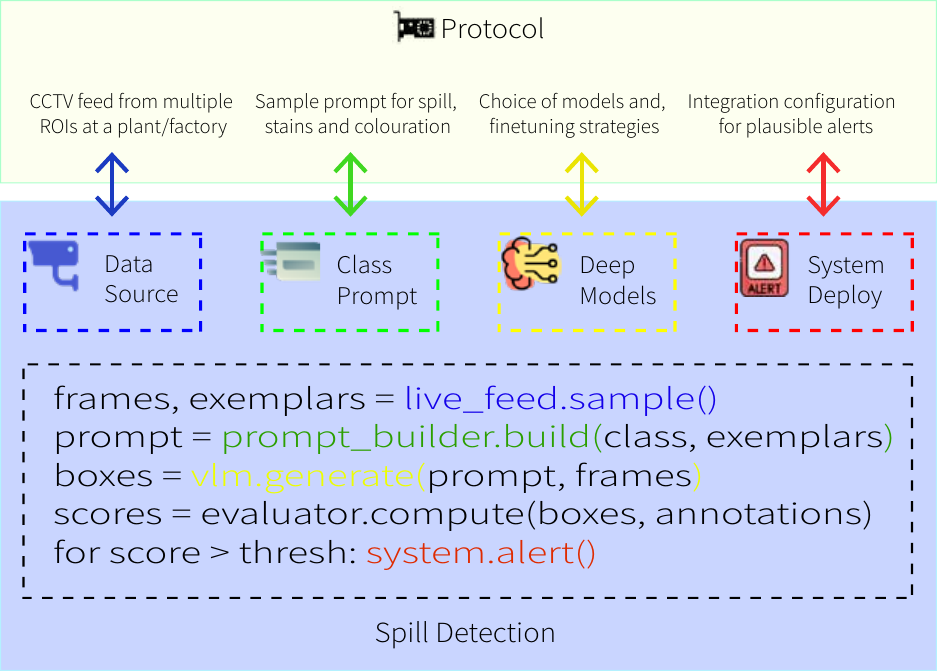}
        \caption{\textbf{Overview of the Industrial Spill Detection Framework.} The system ingests live CCTV feeds alongside a user-defined text prompt. A Vision-Language Model (VLM), selected and fine-tuned via a chosen strategy, analyzes the input to detect and localize potential spills, triggering an alert when the detection confidence surpasses a predefined threshold.}

    \label{fig:pipeline}
\end{figure}

Continuous vigilance is indispensable in industrial operations, where undetected hazards, such as fluid leaks, chemical spills, or mechanical faults, can escalate rapidly, causing economic damage, environmental harm, and risks to human safety~\citep{wang2021survey}. Historically, industrial monitoring has relied on manual inspections or fixed physical sensors. However, both approaches have limitations: human inspections are error-prone and infeasible at scale, while static sensors offer limited spatial awareness, they may indicate \emph{that} a hazard exists, but not \emph{where} or \emph{what} it is.

Computer vision has emerged as a promising solution, offering automated, context-aware surveillance. Modern object detectors such as YOLO, and DETR~\citep{hidayatullah2025yolov8yolo11comprehensivearchitecture, sapkota2025rfdetrobjectdetectionvs, zhang2022dinodetrimproveddenoising, yolov8, jegham2025yoloevolutioncomprehensivebenchmark} push the frontier in detection speed and accuracy. These models perform remarkably well in structured environments, but their real-world deployment in industrial settings remains bottlenecked by data scarcity. Industrial incident data is rare, often proprietary, and difficult to annotate, making it infeasible to collect large-scale, diverse datasets. Consequently, even state-of-the-art detectors tend to overfit to the domains they are trained on, struggling to generalize across variable lighting, or facility layouts~\citep{wang2021survey, Saleh_2021}.

To address these limitations, research has explored multiple paths. Classical anomaly detection methods, such as Gaussian pyramid differencing~\citep{adams1995pyramid}, offer fast but semantically blind solutions. More recently, Vision-Language Models (VLMs)~\citep{jia2021scalingvisualvisionlanguagerepresentation, radford2021learningtransferablevisualmodels} have shown promise for zero-shot recognition in open-world settings. While pretrained VLMs excel at broad generalization, their localization ability and understanding of domain-specific visual cues remain limited without task adaptation.

We present a framework that addresses the core data bottleneck through generative AI and Parameter-Efficient Fine-Tuning (PEFT). Instead of relying on scarce real-world data, we develop a high-fidelity synthetic data generation pipeline that combines Stable Diffusion XL, IP adapters, and inpainting to simulate a wide range of realistic spills.

This synthetic dataset is then used to adapt VLMs via PEFT strategies like Low-Rank Adaptation (LoRA)~\citep{hu2022lora}, enabling the incorporation of domain expertise by updating only a small fraction of model weights.

Importantly, we show that this synthetic corpus benefits \textit{both} vision-language and object detection models. Our experiments reveal that fine-tuning RF-DETR and YOLOv11 on the synthetic+web corpus substantially improves their real-world detection accuracy. However, in low-data regimes where no synthetic data is available, VLMs exhibit stronger generalization than these task-specific detectors. Once synthetic data is introduced, both approaches become competitive, underscoring the broad utility of our pipeline.

\noindent Our contributions in this work include:
\begin{enumerate}
    \item Evaluation of vision-language models for spill detection in industrial settings. We implemented few-shot approaches such as In-Context Learning (ICL) and Parameter-Efficient Fine-Tuning (PEFT) using LoRA for domain adaptation.
    \item Introduction of a novel, controllable pipeline \textit{AnomalInfusion} for generating diverse and photorealistic industrial spill imagery using guided Stable Diffusion with IP adapters and anomaly inpainting.
    \item Demonstration that our synthetic dataset improves performance across both VLMs and state-of-the-art object detectors (YOLOv11, RF-DETR Base throughout the paper), highlighting its broad applicability and critical importance in safety-critical, data-scarce domains.
\end{enumerate}

%% file: sec/2_relatedwork.tex
\section{Related Work}
\label{sec:relatedwork}

The pursuit of automated industrial hazard detection has evolved across three interlinked research frontiers: (i) the progression from supervised detectors to vision-language foundation models, (ii) the use of synthetic data to overcome annotation bottlenecks, and (iii) the emergence of parameter-efficient fine-tuning (PEFT) techniques for scalable domain adaptation. Our work situates itself at the intersection of these areas.

\begin{figure*}[t]
    \centering
    \begin{subfigure}[c]{0.40\linewidth}
        \centering
        \includegraphics[width=\linewidth]{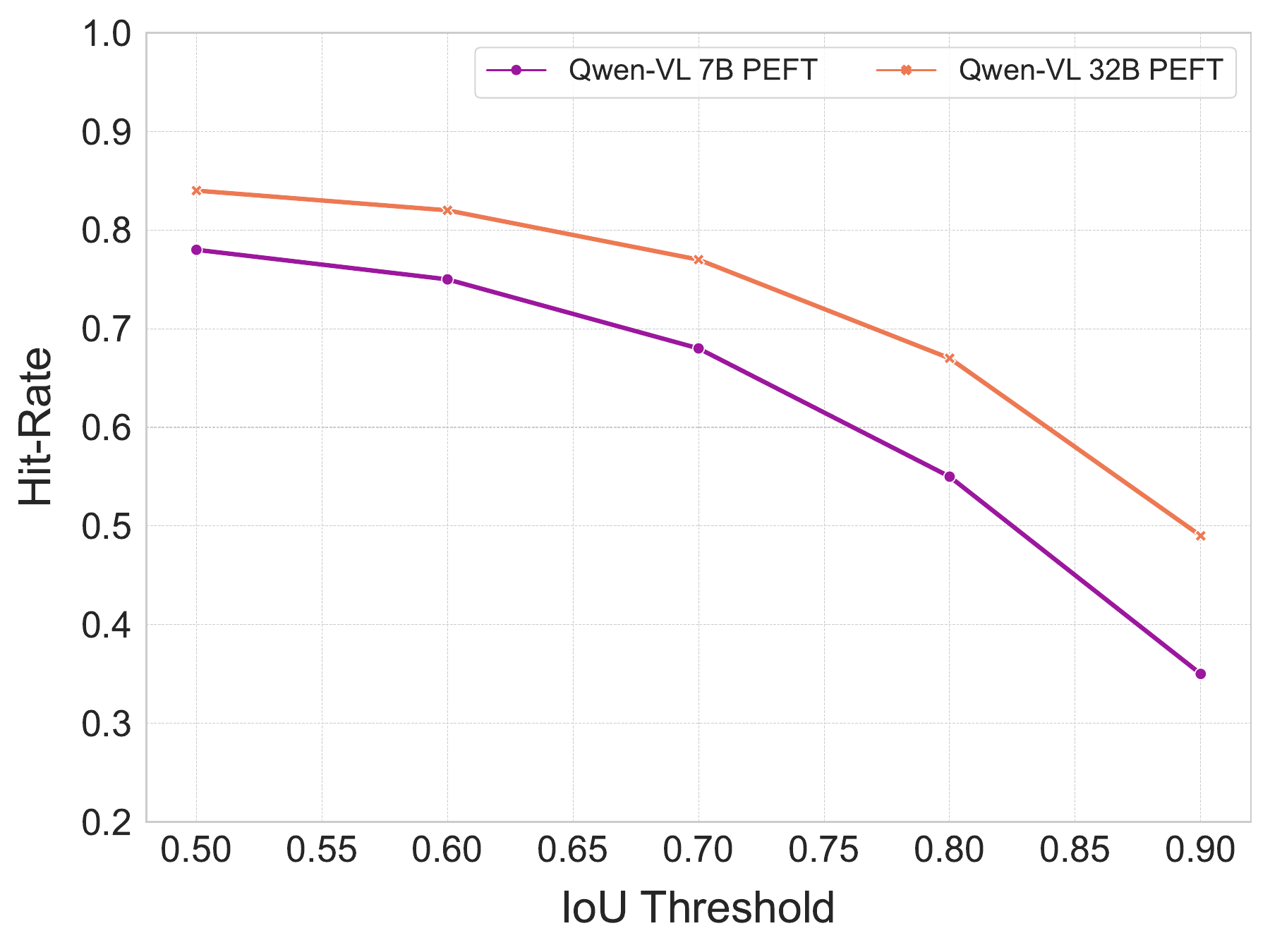}
        \caption{Localization accuracy as a function of IoU threshold.}
        \label{fig:localization_accuracy}
    \end{subfigure}
    \hfill
    \begin{subfigure}[c]{0.55\textwidth}
        \centering
        \includegraphics[width=\linewidth]{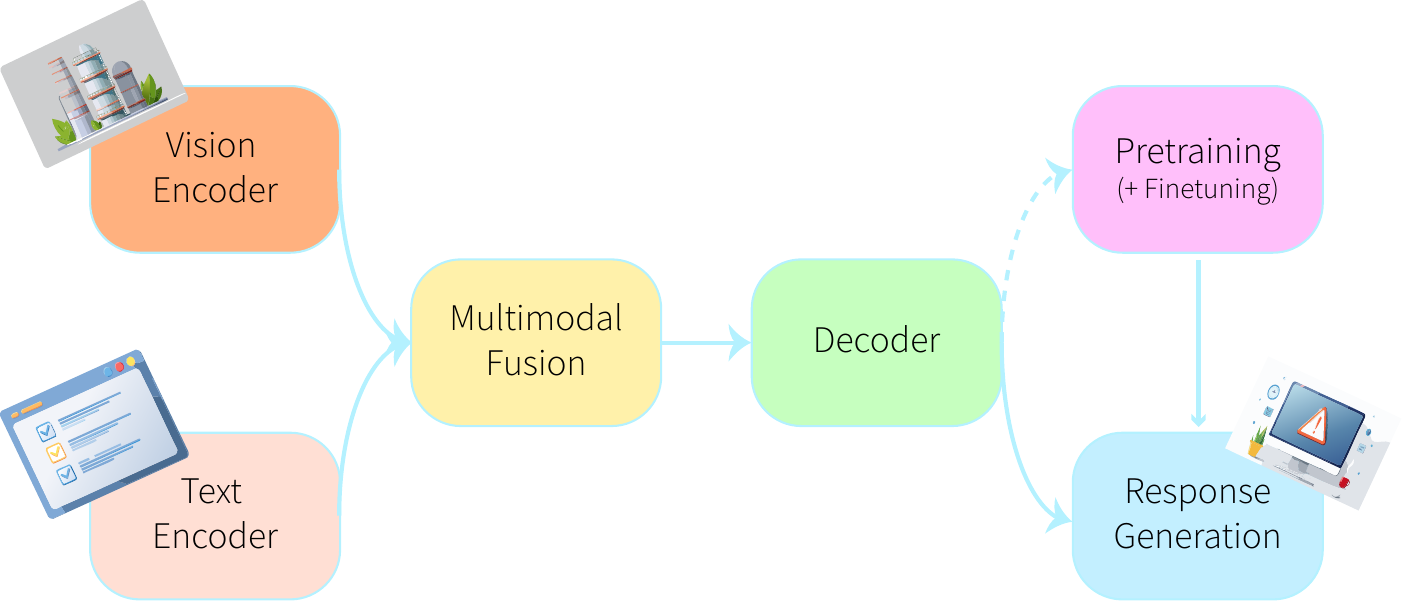}
        \caption{High Level Overview of PEFT Flow}
    \end{subfigure}
    \caption{\textbf{Adaptation strategies and their impact on spatial precision.} The left subfigure illustrates that LoRA-adapted models retain higher localization accuracy at stricter IoU thresholds, validating the effectiveness of synthetic supervision. The right subfigure presents a component-level breakdown of LoRA application across vision and language backbones.}
    \label{fig:adaptation_comparison}
\end{figure*}

\noindent\textbf{From Supervised Detectors to Foundation Models.}
Traditional industrial visual monitoring systems began with classical techniques such as background subtraction and image differencing~\citep{radke2005image}, which were computationally lightweight but highly sensitive to environmental changes. The advent of deep learning introduced powerful supervised object detectors, including two-stage approaches like Faster R-CNN~\citep{ren2015faster} and one-stage detectors such as YOLOv3~\citep{redmon2018yolov3} and DETR~\citep{carion2020endtoendobjectdetectiontransformers}. More recent successors, such as YOLOv7--v12~\citep{wang2022yolov7trainablebagoffreebiessets,hidayatullah2025yolov8yolo11comprehensivearchitecture} DETR, DINO, SAM and RF-DETR~\citep{zhang2022dinodetrimproveddenoising, liu2023grounding, kirillov2023segany, sapkota2025rfdetrobjectdetectionvs}, have pushed state-of-the-art performance in real-time detection and long-tail robustness. Yet, these models depend heavily on large-scale annotated datasets, which are unavailable in sensitive or safety-critical environments.

To address generalization under data scarcity, recent efforts have turned toward \textbf{Vision-Language Models (VLMs)} such as CLIP~\citep{radford2021learningtransferablevisualmodels}, ALIGN~\citep{jia2021scalingvisualvisionlanguagerepresentation}, and Florence~\citep{yuan2021florencenewfoundationmodel}, which are pre-trained on web-scale image-text pairs. These models exhibit strong zero-shot transfer for image-level classification and visual grounding. Building upon this, open-vocabulary detectors like GLIP~\citep{li2022grounded}, GroundingDINO~\citep{liu2023grounding}, and Grounded-SAM~\citep{kirillov2023segment} combine detection and segmentation under weak supervision. However, these systems still degrade in industrial environments due to severe domain shifts, uncommon textures, lighting conditions, and anomaly types that are underrepresented in pretraining corpora.

\noindent\textbf{Synthetic Data and the Sim-to-Real Gap.}
Synthetic data has emerged as a practical solution to the annotation bottleneck. Earlier approaches used game engines or 3D simulators (e.g., CARLA~\citep{dosovitskiy2017carlaopenurbandriving}, AI2THOR~\citep{kolve2022ai2thorinteractive3denvironment}) to render synthetic environments for training. However, limited realism in such renderings often introduced a "sim-to-real" gap~\citep{chen2022understandingdomainrandomizationsimtoreal}. Diffusion models~\citep{ho2020denoisingdiffusionprobabilisticmodels} have revolutionized this space, enabling high-fidelity generation of semantically aligned imagery. Text-to-image models like DALLE-2~\citep{ramesh2022hierarchicaltextconditionalimagegeneration} and Stable Diffusion~\citep{rombach2022highresolutionimagesynthesislatent} now support conditioning on detailed prompts or images via ControlNet~\citep{zhang2023addingconditionalcontroltexttoimage}, IP-Adapters~\citep{ye2023ipadaptertextcompatibleimage}, or DreamBooth~\citep{ruiz2023dreamboothfinetuningtexttoimage}, improving visual grounding in generations.

Several works have demonstrated the efficacy of synthetic imagery for detection and segmentation: GenAug~\citep{hendrycks2022pixmixdreamlikepicturescomprehensively}, Domain Randomization~\citep{chen2022understandingdomainrandomizationsimtoreal}, Task2Sim~\citep{mishra2022task2simeffectivepretraining}, and DreamFusion~\citep{poole2022dreamfusiontextto3dusing2d}. Others, such as StyleGAN-based simulators~\citep{karras2020analyzingimprovingimagequality,liu2024bestpracticeslessonslearned}, have been used for compositional domain transfer. Yet, most existing works focus on \textit{training small models} or \textit{augmenting real datasets}, rather than fully adapting \textit{large foundation models} using synthetic data alone in domains with zero or near-zero real samples. Applications in industrial anomaly detection remain underexplored despite recent interest~\citep{moenck2024industriallanguageimagedatasetilid}.

\noindent\textbf{Efficient Domain Adaptation via PEFT.}
Full fine-tuning of foundation models is computationally expensive and prone to catastrophic forgetting~\citep{Kirkpatrick_2017}. To address this, Parameter-Efficient Fine-Tuning (PEFT) has emerged as a compelling alternative. LoRA~\citep{hu2022lora}, QLoRA~\citep{dettmers2023qlora}, AdapterFusion~\citep{pfeiffer2021adapterfusionnondestructivetaskcomposition}, and BitFit~\citep{zaken2022bitfitsimpleparameterefficientfinetuning} allow for tuning less than 1\% of a model’s weights while retaining strong downstream performance. While widely adopted in NLP, vision applications of PEFT remain nascent. Recent works like VPT~\citep{jia2022vpt}, SSF~\citep{lian2023scalingshiftingfeatures}, and AdaptFormer~\citep{chen2022adaptformeradaptingvisiontransformers} apply PEFT to vision transformers, but few zero-shot detection regimes or examine PEFT’s behavior under extreme data scarcity.

Our work advances the field by synthesizing these three threads. We present a unified framework that: (1) leverages high-fidelity diffusion-based generation to create an industrial spill dataset without real incident data, (2) adapts VLMs using parameter-efficient strategies like LoRA, and (3) shows that the resulting models are not only competitive with fully fine-tuned detectors but also generalize better in low-data or zero-shot regimes. To our knowledge, this is the first systematic study of adapting foundation vision-language models for industrial hazard detection using purely synthetic data, bridging the gap between generalist pretraining and specialized safety-critical deployment.

%% file: sec/3_setup.tex
\section{Experimental Setup}
\label{sec:experimental}

Designing a study that speaks simultaneously to plant-floor practitioners and AI researchers requires a careful balance: the system must be reproducible with modest resources, yet documented with sufficient technical detail to withstand scrutiny. Our experiments evaluate how Vision-Language Models (VLMs) can be adapted for industrial spill detection using synthetic and publicly available data.

\subsection{Vision Language Models}
\label{sec:vlms}

VLMs align image and text modalities through joint pretraining on large-scale image–caption datasets~\citep{radford2021learningtransferablevisualmodels,jia2021scalingvisualvisionlanguagerepresentation}. At inference, they process an image and a text prompt (e.g., ``Where is the spill?'') to produce either textual answers or structured outputs such as bounding boxes.

We adopt the open-source \textbf{Qwen2.5-VL} ~\citep{bai2025qwen25vltechnicalreport} family at three parameter scales: 3B, 7B, and 32B. These models combine a Swin-style vision transformer~\citep{liu2021swintransformerhierarchicalvision} with a causal language decoder to form a shared vision-language latent space. The 3B model is suitable for single-GPU settings, while the 32B variant offers greater grounding fidelity.

\subsection{Structured Prompting}
\label{sec:prompting}

Each image is paired with a structured prompt designed to elicit precise and conservative detection behavior in high-stakes environments. The system prompt simulates the reasoning of an industrial inspector, while the user prompt requests bounding-box outputs in YOLO v11/COCO JSON format for one of eight anomaly classes (e.g., \texttt{oil-spill}, \texttt{floor-stain}, \texttt{chemical-discoloration}):\\

\noindent \textbf{System:} \textit{"You are a certified industrial safety inspector specializing in hazardous spill, leak, and stain detection across factories and energy plants. Only report verifiable safety hazards. Do not guess or speculate."}\\
\noindent \textbf{User:} \textit{"Detect and return the bounding-box coordinates of the \textless class\textgreater\ in COCO JSON format, if present."}

To encourage concise and reliable outputs, we set the decoding parameters as follows: temperature $\tau = 0.10$, nucleus sampling $p = 0.001$, and a repetition penalty of 1.2.

\subsection{Parameter-Efficient Fine-Tuning (PEFT)}
\label{sec:peft}

We evaluate three adaptation strategies to assess performance under varying resource and data constraints:

\paragraph{Zero-Shot Inference.}
The model is used as-is without any additional tuning. This baseline assesses the pretrained VLM’s out-of-the-box generalization to spill detection.
\vspace{-0.5em}
\paragraph{In-Context Learning (ICL).}
Prepend $k \in \{5,10,15\}$ labeled examples (image + bounding box) to input prompt~\citep{brown2020language,lin2025generalizationenhancedfewshotobjectdetection, zhou2024visualincontextlearninglarge}. The model weights remain frozen. The prediction is conditioned on a support set $S = \{(\mathbf{x}_i, y_i)\}_{i=1}^{k}$:
\begin{equation}
    \mathbb{P}(y \mid \mathbf{x}, S),
\end{equation}
Utilizing few solved examples available at inference time.
\vspace{-0.5em}
\paragraph{Low-Rank Adaptation (LoRA).}
LoRA~\citep{hu2022lora} introduces lightweight trainable adapters that inject domain-specific knowledge by modifying only a small number of parameters. Instead of fine-tuning the full weight matrix $W \in \mathbb{R}^{d_\text{out} \times d_\text{in}}$, LoRA learns two low-rank matrices $A$ and $B$:
\begin{equation}
    W' = W + \alpha AB.
\end{equation}
Where $A \in \mathbb{R}^{d_\text{out} \times r}, \quad B \in \mathbb{R}^{r \times d_\text{in}}, \quad \alpha = \frac{1}{r}$.\\
We evaluate three LoRA configurations:
\begin{itemize}
    \item \textbf{LoRA-L:} Language pathway adapted only.
    \item \textbf{LoRA-V:} Vision encoder adapted only.
    \item \textbf{LoRA-(V+L):} Both pathways adapted jointly.
\end{itemize}
\noindent For all configurations and Datasets LoRA is trained for 800 steps (with early stopping) with a batch size of 8 and 4 step gradient accumulation (effective batch size 32). Rank is fixed to 8 for our experiments with LR set to $5*10^{-5}$.
\subsection{Datasets}
\label{sec:datasets}

\begin{figure}[t]
    \centering
    \includegraphics[width=\linewidth]{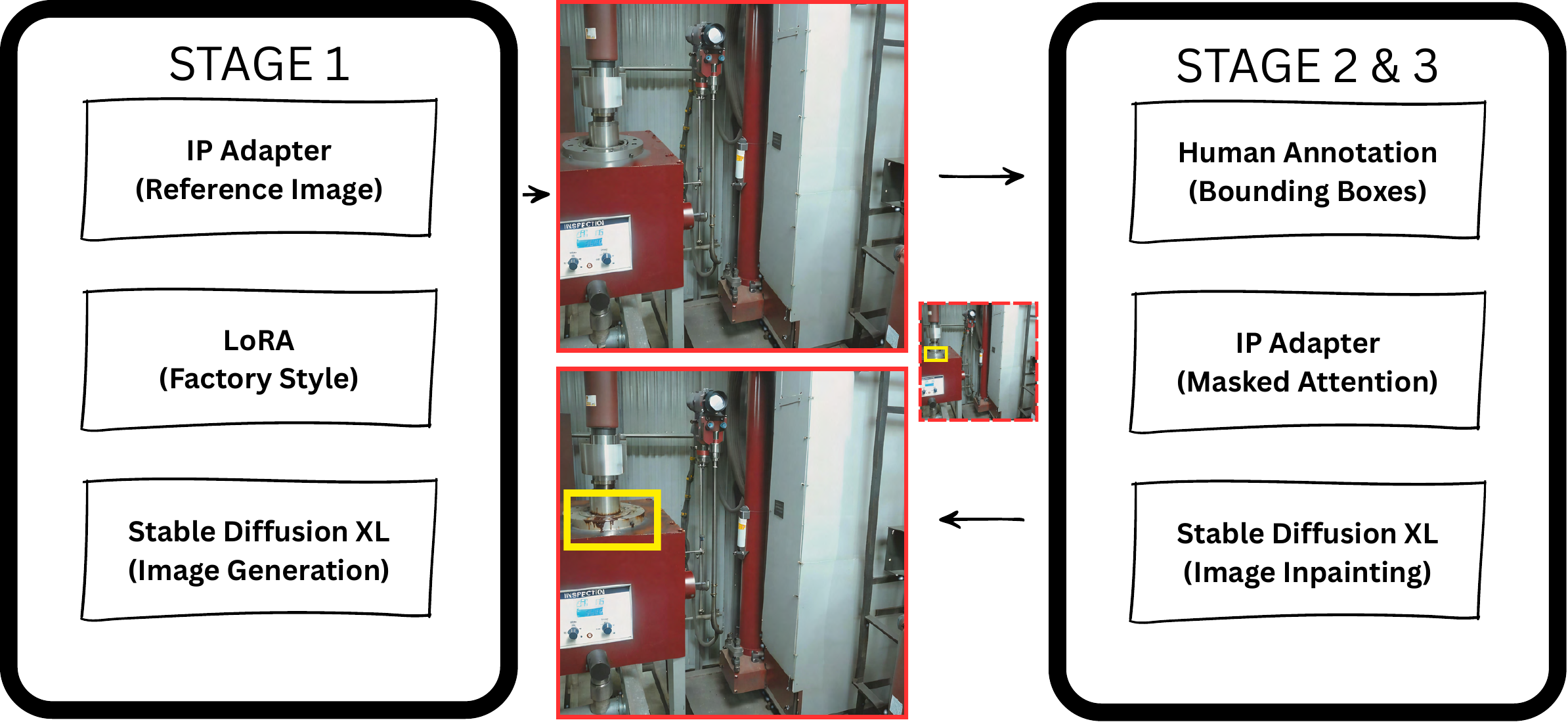}
    \caption{End-to-end synthetic data generation workflow. Stage 1 constructs factory-style backgrounds guided by reference images and structural maps via Stable Diffusion XL~\citep{rombach2022highresolutionimagesynthesislatent}, IP Adapter~\citep{ye2023ipadaptertextcompatibleimage}, and LoRA~\citep{luo2023lcmlorauniversalstablediffusionacceleration}. Stages 2 and 3 add hazards: bounding boxes are manually placed in plausible locations, and inpainting is performed using SDXL with spill-specific prompts and conditioning.}
    \label{fig:se_det_synth}
\end{figure}

We evaluate on two real-world datasets, a fraction of these is held out from training and generation phases to ensure a rigorous test of sim-to-real generalization:

\begin{itemize}
    \item \textbf{Public  Spill Data:} A web scraped dataset sourced from \href{https://roboflow.com/}{Roboflow} public datasets and thoroughly deduplicated. The sources are carefully chosen to keep it as close to our Factory Site spills as possible. 1520 images collected out of which 520 are used for evaluation and testing and the rest 100 are used for PEFT and FT of VLMs and SOTA Object Detectors respectively. 
    \item \textbf{Proprietary Factory Data:} A challenging, in-house dataset collected from operational industrial sites, featuring subtle leaks, stains, and occlusions in complex environments. 150 images in total out of which 50 are separated for ICL and remaining 100 are used for Evaluation.
    \item \textbf{Synthetic Spill (SynSpill)  Dataset:} A curated dataset of 2000 photorealistic synthetic spill images generated via our AnomalInfusion pipeline (see Section~\ref{sec:synth_data}), all of it is used for PEFT and FT purposes.
\end{itemize}

\subsection{Evaluation Metric}
\label{sec:metrics}

We report the \textbf{mean hit rate} at an IoU threshold of 0.5 across all anomaly classes. A predicted bbox is counted as a \textit{hit} if it has sufficient overlap with the ground truth:

\begin{equation}
    \text{IoU} = \frac{|B_{\text{pred}} \cap B_{\text{gt}}|}{|B_{\text{pred}} \cup B_{\text{gt}}|}, \quad \text{Hit if } \text{IoU} \ge 0.5.
\end{equation}
This captures detection accuracy and spatial precision, aligning with requirements in safety-critical monitoring.

%% file: sec/4_synthetic.tex
\section{Synthetic Data Generation}
\label{sec:synth_data}

While supervised learning and fine-tuning have enabled significant advances in vision systems, their success ultimately hinges on the availability of labeled data. In safety-critical domains like industrial spill detection, this presents a core challenge: hazardous incidents are rare, sensitive, and difficult to capture due to privacy, safety, and operational constraints. Consequently, models must learn to detect anomalies they may never encounter in real-world training data.

We address this paradox by recognizing that while hazardous \textit{events} are rare, their visual \textit{signatures} can be synthesized. We introduce a scalable, three-stage synthetic data pipeline that generates photorealistic, structurally plausible spill scenarios anchored in real industrial imagery. Using only a small seed set of 150 unlabelled factory images as style references, our method produces a corpus of 2,000 synthetic images with precise annotations, designed to serve as a drop-in adaptation dataset for hazard detection.

\subsection*{Stage 1: Domain-Anchored Scene Generation}

The foundation of our pipeline is the generation of diverse, context-aware background images. We employ a \textbf{triple-guided generative process} using Stable Diffusion XL (SDXL) to ensure outputs are both photorealistic and domain-consistent. The three guidance signals are: 

\noindent \textbf{Textual Prompts:} Define semantic content, architectural elements (e.g., \textit{concrete floor}, \textit{metal catwalk}), lighting (\textit{diffuse fluorescent lighting}), and scene types (\textit{control room}, \textit{factory corridor}). \textbf{IP-Adapter Conditioning:} Transfers stylistic priors from 150 real factory images, grounding the generation in authentic textures, colors, and spatial layouts. \textbf{LoRA Style Injection:} Applies low-rank adaptation modules to reinforce industrial realism and consistency.

Together, these mechanisms produce a diverse set of ``clean'' factory scenes that are not merely synthetic, but visually faithful to operational plant environments.

\begin{figure*}[t]
    \centering
    \begin{subfigure}[c]{0.65\textwidth}
        \centering
        \includegraphics[width=\linewidth]{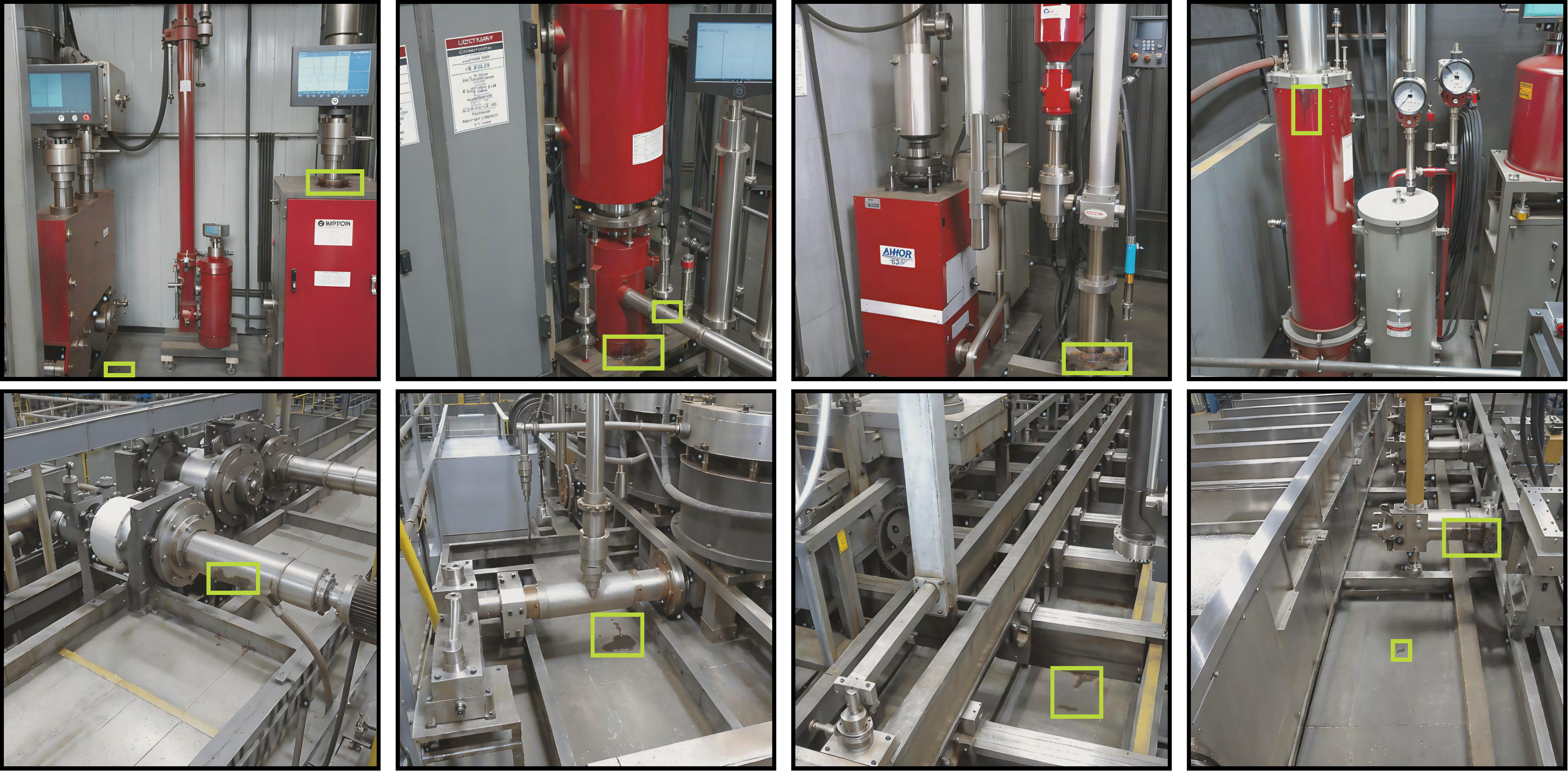}
        \caption{Representative synthetic image samples generated using triple-guided diffusion.}
    \end{subfigure}
    \hfill
    \begin{subfigure}[c]{0.30\textwidth}
        \centering
        \includegraphics[width=\linewidth]{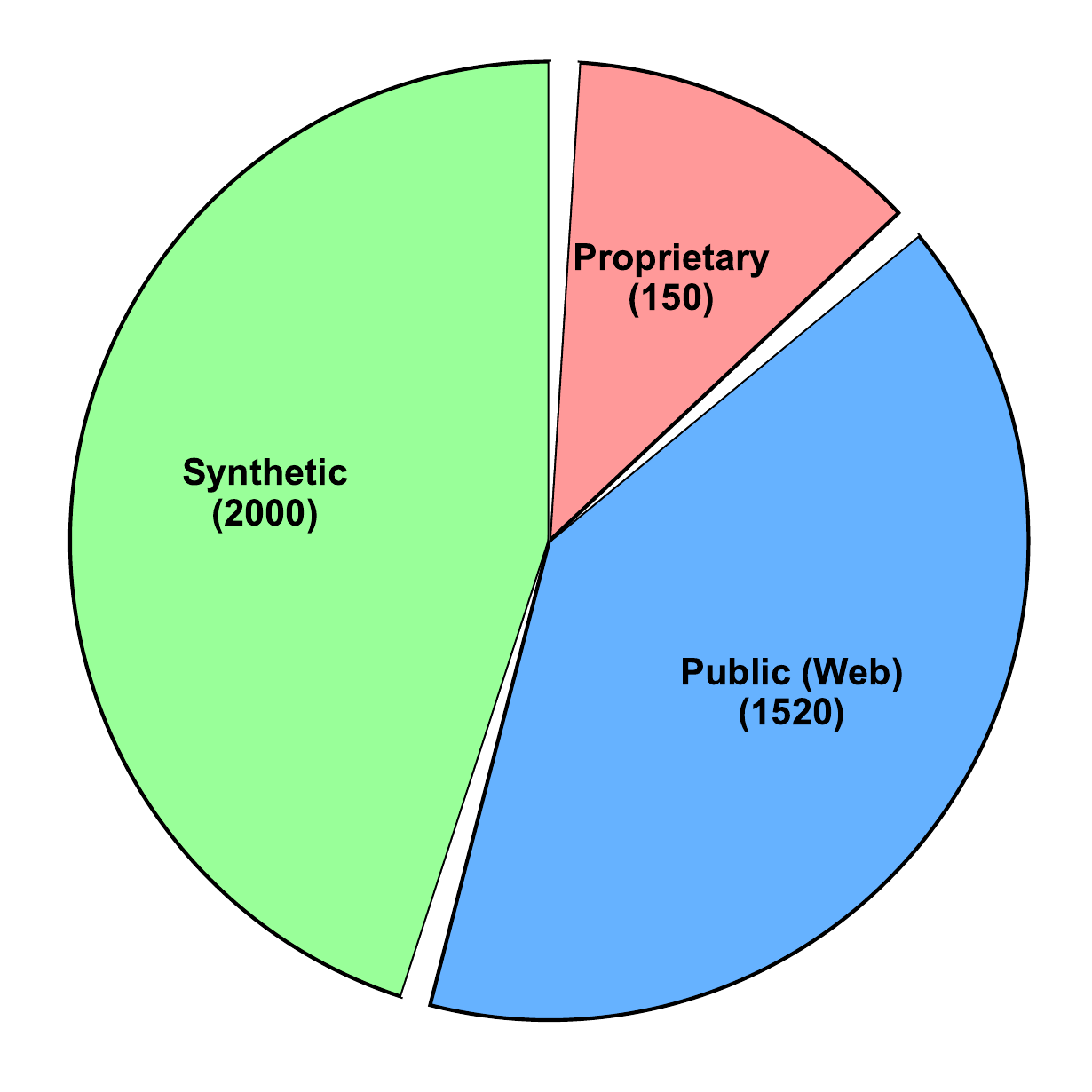}
        \caption{Distribution across sources: synthetic (2,000), public (1,520), and proprietary (150).}
    \end{subfigure}
    \caption{\textbf{Dataset composition for adaptation and evaluation.} The left panel highlights the variability achieved through synthetic generation, while the right panel summarizes the total image count from each source category.}
    \label{fig:data_overview}
\end{figure*}

\begin{table}[ht!]
    \centering
    \caption{Key hyperparameters used in synthetic image generation.}
    \label{tab:gen_params}
    \resizebox{0.9\linewidth}{!}{
    \begin{tabular}{@{}ll@{}}
        \toprule
        \textbf{Parameter} & \textbf{Value / Configuration} \\
        \midrule
        \multicolumn{2}{@{}l}{\textit{\textbf{Scene Generation Specifics}}} \\
        Base Model & Stable Diffusion XL 1.0~\citep{interiorSceneXL2025} \\
        Image Resolution & 1024 \texttimes{} 1024 \\
        Sampler & DDPM-SDE-2m-GPU \\
        Scheduler & Karras \\
        Sampling Steps & 64 \\
        CFG Scale & 8 \\
        LoRA Strength & 0.2–0.4 \\
        IP-Adapter & IP Composition+CLIP-ViT-H~\cite{laion_clip_vith14_laion2b_2023,ostris_ipcompositionadapter_2024}\\
        IP-Adapter Strength & 0.6 \\
        \midrule
        \multicolumn{2}{@{}l}{\textit{\textbf{Inpainting Specifics}}} \\
        Inpainting Model & SDXL-Turbo Inpainting~\citep{stabilityai_sdxl-turbo_fp16_2023} \\
        Differential Diffusion & Enabled \\
        Mask Feathering & 50 pixels \\
        Mask Opacity & 75\%\\
        Denoise Strength & 0.5-0.6 \\
        \bottomrule
    \end{tabular}
    }
\end{table}

\subsection*{Stage 2: Expert-Guided Anomaly Localization}

A key novelty of our pipeline lies in how anomalies are introduced. Rather than inserting spills arbitrarily, which risks generating unrealistic samples, we use a \textbf{human-in-the-loop annotation step}. An experienced annotator identifies semantically appropriate regions for hazards: valves, pipe junctions, or near frequently serviced equipment.

Each bounding box encodes an implicit causal logic (“a valve may leak here”), aligning with how spills manifest in real scenarios. This step ensures our corpus reflects operational plausibility rather than synthetic randomness.

\subsection*{Stage 3: Physically-Plausible Spill Inpainting}

Given the annotated regions, we simulate hazards via \textbf{differential inpainting}~\citep{lugmayr2022repaintinpaintingusingdenoising,levin2024differentialdiffusiongivingpixel}. A soft binary mask is generated for each bounding box and passed to SDXL’s inpainting mode. This allows the anomaly to blend seamlessly with the surrounding scene. We apply layered conditioning to preserve fidelity and coherence:
\begin{itemize}
    \item \textbf{Prompts} Material Description type (e.g., \textit{glistening oil}).
    \item \textbf{IP-Adapter} references real-world spill textures.
    \item \textbf{LoRA} maintains lighting, surface, style and texture.
\end{itemize}

\noindent Spills are visually realistic in isolation and contextually integrated into the background, respecting physical constraints and visual consistency (Figures in Supplementary).

\subsection*{Pipeline Summary and Implementation Details}

The full pipeline, background generation, expert-informed bounding box annotation, and guided inpainting, is lightweight, modular, and scalable. Table~\ref{tab:gen_params} summarizes the core generation parameters used at each stage.

\subsection*{Distribution and Dataset Insights}

The Synthetic corpus spans diverse lighting, geometry, spill morphology, and environmental context, closely mirroring the operational variability of real-world industrial sites. Annotations are exported in COCO format and are used to Inpaint the images with spills using the same bbox coordinates for feathered mask generation, following that the generated images with annotations are directly used to train both VLMs and object detectors. We also aggregate 1,520 additional web-scraped images from publicly available industrial datasets to broaden test diversity and conduct generalization studies. Combined with our 150-image proprietary set, these form a rich evaluation suite.

Unlike prior work that either (i) generates fully synthetic environments, or (ii) overlays textures without semantic guidance, our method offers a principled alternative by bridging the sim-to-real gap with tailored generation and expert-in-the-loop realism checks, we create a dataset that enables robust adaptation in domains where real data is scarce, sensitive, or unavailable.

This synthetic corpus forms the foundation for the experiments described next, where we show that models adapted on this data not only outperform their zero-shot counterparts, but in many cases approach or exceed the performance of detectors trained on curated real-world subsets.

%% file: sec/5_experiments.tex
\section{Results and Discussion}
\label{sec:experiments}

We organize our findings in two phases. First, we evaluate baseline performance of Vision-Language Models (VLMs) under few-shot and parameter-efficient tuning regimes using limited real-world data. Then, we assess the full proposed pipeline VLM adaptation using high-fidelity synthetic data and benchmark it against SOTA object detectors.

\begin{figure*}[t]
    \centering
    \begin{subfigure}{0.98\textwidth}
        \centering
        \includegraphics[width=\linewidth]{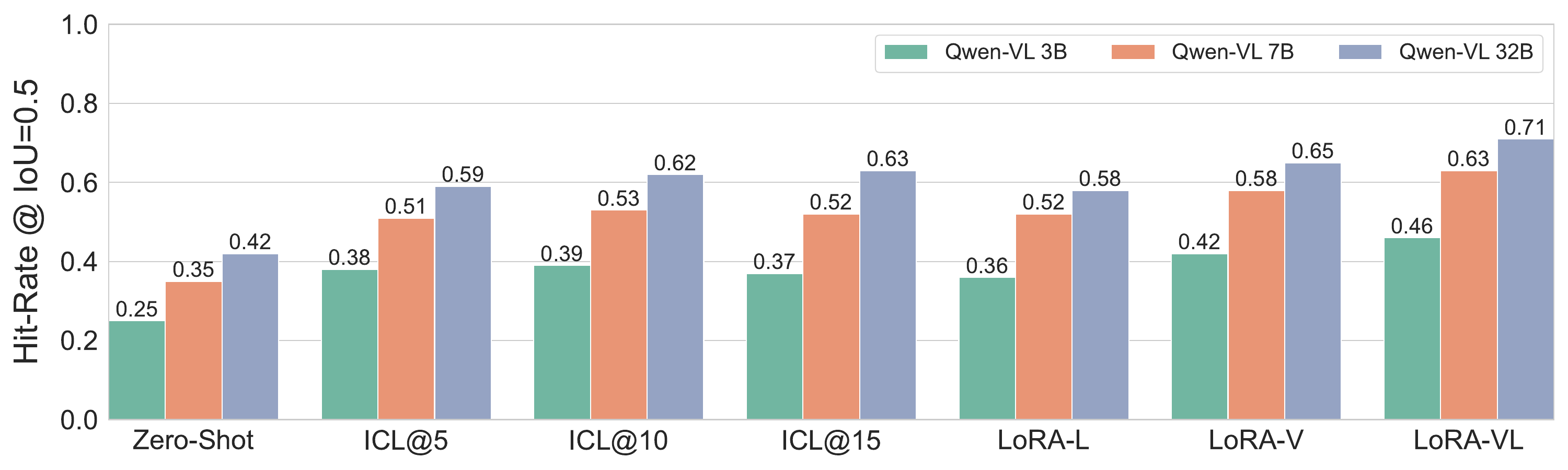}
        \caption{Performance of different size variants of Qwen 2.5 VL across Performance Enhancing Strategies.}
    \end{subfigure}
    
    \vspace{0.5em} % optional spacing between the subfigures
    \begin{subfigure}{0.98\textwidth}
    \centering
        \includegraphics[width=\linewidth]{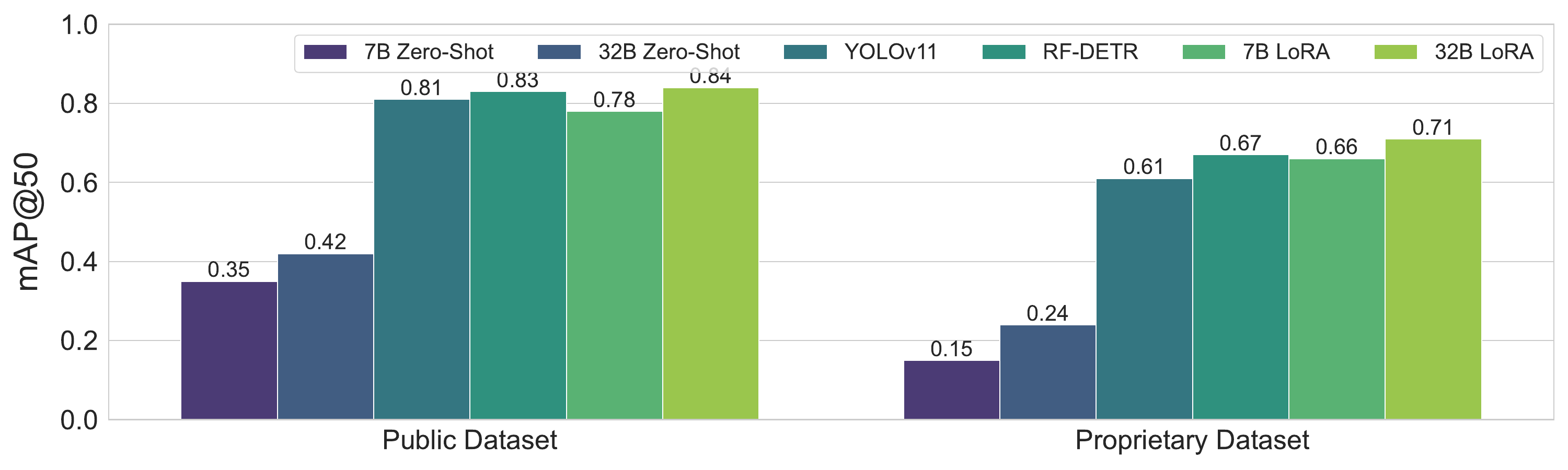}
        \caption{Overall detection performance across different model types and adaptation strategies.}
    \end{subfigure}

    \caption{Performances across Methods, Models, and Datasets}
    \label{fig:performance}
\end{figure*}

\vspace{0.5em}
\begin{table}[hbt!]
    \centering
    \caption{Mean hit-rate @ IoU = 0.5 on Public Evaluation Set.}
    \label{tab:overall}
    \begin{tabular}{lccc}
        \toprule
        \textbf{Method} & \textbf{3B} & \textbf{7B} & \textbf{32B} \\
        \midrule
        Zero-Shot       & 0.25 & 0.35 & 0.42 \\
        ICL (5 shots)   & 0.38 & 0.51 & 0.59 \\
        ICL (10 shots)  & 0.39 & 0.53 & 0.62 \\
        ICL (15 shots)  & 0.37 & 0.52 & 0.63 \\
        LoRA (L)        & 0.36 & 0.52 & 0.58 \\
        LoRA (V)        & 0.42 & 0.58 & 0.65 \\
        LoRA (V+L)      & \textbf{0.46} & \textbf{0.63} & \textbf{0.71} \\
        \bottomrule
    \end{tabular}
\end{table}

\vspace{-0.5em}
\begin{table}[hbt!]
    \centering
    \caption{Mean hit-rate @ IoU = 0.5 on Proprietary Evaluation Set.}
    \label{tab:hit_rates_se}
    \begin{tabular}{lccc}
        \toprule
        \textbf{Method} & \textbf{3B} & \textbf{7B} & \textbf{32B} \\
        \midrule
        Zero-Shot       & 0.11 & 0.15 & 0.24 \\ 
        ICL (5 shots)   & 0.21 & 0.26 & 0.33 \\
        ICL (10 shots)  & 0.24 & 0.29 & 0.34 \\
        ICL (15 shots)  & 0.23 & 0.28 & 0.36 \\
        LoRA (L)        & 0.19 & 0.23 & 0.31 \\
        LoRA (V)        & 0.26 & 0.31 & 0.41 \\
        LoRA (V+L)      & \textbf{0.29} & \textbf{0.34} & \textbf{0.49} \\
        \bottomrule
    \end{tabular}
\end{table}

\subsection{Adaptation Under Real-World Data Scarcity}

\textbf{Setup:}  
We assess three adaptation strategies on Qwen2.5-VL models of different sizes (3B, 7B, 32B): Zero-Shot Inference, In-Context Learning (ICL), and LoRA-based Parameter-Efficient Fine-Tuning (PEFT).
Performance is measured by mean hit-rate at IoU $\geq$ 0.5 on both public and proprietary datasets. \textbf{Few-shot Learning: Promise and Limits:}  
ICL with 5 examples improves performance significantly, boosting the 7B model’s hit rate by over 45\% compared to its zero-shot baseline. However, benefits plateau quickly. Adding more shots (10 or 15) provides only marginal gains or occasional regressions. This aligns with prior work highlighting ICL’s brittleness under token limitations~\citep{chen2024iclevalevaluatingincontextlearning}. \textbf{Model Scale vs. Effective Adaptation:}  
Larger models consistently perform better, but raw scale is not enough. For instance, a 3B model fine-tuned with LoRA-(V+L) rivals or surpasses the 7B model in zero-shot or ICL settings. This underscores that \textit{domain adaptation, not just parameter count, is essential to unlock model capacity}. \textbf{Vision Matters Most.}  
LoRA-V (vision-only tuning) consistently outperforms LoRA-L (language-only), revealing that visual specificity, textures, shadows, gradients, plays a more critical role than textual understanding in this domain. LoRA-(V+L) offers additional gains through joint adaptation. \textbf{Improved Recognition = Tighter Localization:}  
As shown in Figure~\ref{fig:localization_accuracy}, improvements in detection correlate with higher spatial accuracy. Adapted models maintain better performance even at stricter IoU thresholds, validating the quality of supervision. \textbf{Adaptation Gain vs. Domain Gap:}  
We quantify relative improvement using:
\[
\Delta_m = \left( \frac{\text{HR}_m - \text{HR}_{\text{ZS}}}{\text{HR}_{\text{ZS}}} \right) \times 100\%.
\]
For Qwen2.5-VL-7B on the public set, LoRA-(V+L) yields an uplift of 80\%. However, on the proprietary set, the same configuration drops to 0.34, underscoring that even parameter-efficient tuning is ultimately constrained by the coverage of available data. \textbf{Insight:} VLMs generalize locally (to seen textures or lighting) but not structurally. In safety-critical domains, this is insufficient.

\begin{table*}[ht!]
    \centering
    \caption{Main performance comparison (mAP@50).}
    \label{tab:main_results_updated}
    \begin{tabular}{lcc}
        \toprule
        \textbf{Model / Method} & \textbf{Public Dataset (mAP@50)} & \textbf{Proprietary (mAP@50)} \\
        \midrule
        Qwen-VL 7B (Zero-Shot) & 0.35 & 0.15 \\
        Qwen-VL 32B (Zero-Shot) & 0.42 & 0.24 \\
        \midrule
        \textit{Baselines (Fine-Tuning w/ Synthetic + Public Data)} \\
        YOLOv11 & 0.81 & 0.64 \\
        RF-DETR & 0.83 & 0.67 \\
        \midrule
        \textit{Proposed Method (PEFT w/ Synthetic + Public Data)} \\
        Qwen-VL 7B + LoRA (V+L) & 0.78 & 0.66 \\
        \textbf{Qwen-VL 32B + LoRA (V+L)} & \textbf{0.84} & \textbf{0.71} \\
        \bottomrule
    \end{tabular}
\end{table*}

\subsection{Adaptation Using Synthetic Data:} To overcome the data ceiling exposed above, we evaluate our full pipeline: adapting Qwen2.5-VL models using synthetic data generated via our AnomalInfusion framework.

\textbf{Setup:}  
We fine-tune Qwen2.5-VL (7B and 32B) using LoRA-(V+L) on our synthetic dataset (2,000 images), and compare to YOLOv11 and RF-DETR baselines trained on the same data. Evaluation uses mAP@50.

\noindent\textbf{Quantitative Results:}
\begin{itemize}
    \item VLMs adapted via PEFT match or exceed detector performance on both datasets.
    \item Qwen2.5-32B + LoRA outperforms RF-DETR by 7 points on the proprietary set.
    \item Zero-shot VLMs lag significantly, reaffirming the need for domain-specific adaptation.
\end{itemize}

\noindent\textbf{Qualitative Trends:}  
As shown in Figure~\ref{fig:data_overview}, PEFT-adapted VLMs are more robust to occlusions, lighting shifts, and background clutter. They detect subtle anomalies, like semi-transparent oil stains, that traditional detectors frequently miss or misclassify.

\subsection{Implications for Real-World Deployment}

\textbf{Fast and Deployable:}  
Our PEFT pipeline runs on a single GPU, adapts models overnight, and requires no end-to-end retraining. It integrates easily with factory CCTV feeds, making it viable for real-world monitoring. \textbf{Towards Predictive Maintenance:} By logging location, time, and severity of detections, the system can feed higher-level analytics for predicting failure points, supporting proactive safety interventions. \textbf{Future-Proof Architecture:} The modular LoRA adaptation strategy ensures compatibility with future foundation models. As new VLMs emerge, our system can adapt with minimal re-engineering.

%% file: sec/6_conclusion.tex
\section{Conclusion}
\label{sec:conclusion}

Industrial hazard detection faces a fundamental barrier: the events we most need to recognize are too rare, sensitive, or dangerous to support large-scale data collection. We address this challenge by pairing high-fidelity synthetic data generation with parameter-efficient adaptation of $VLMs$.

Our three-stage pipeline produces realistic, high-resolution images of industrial spills using just a handful of unlabelled factory references. This yields 2,000 richly annotated scenes that reflect real-world geometry and context, eliminating the need for risky or infeasible data gathering. When used to adapt VLMs via parameter-efficient fine-tuning (PEFT), these synthetic examples lead to strong performance across both public and proprietary test sets, outperforming conventional detectors trained on the same data. This confirms our core insight: foundation models, when paired with carefully designed synthetic data, offer a more effective and scalable path for safety-critical detection than traditional models trained from scratch. SynSpill dataset will be made public upon acceptance.

Crucially, our approach is not only accurate but adaptable. It supports rapid deployment in new settings and seamless updates through data regeneration, without retraining from scratch. This flexibility makes it well-suited for evolving industrial environments.

\noindent In sum, this work provides a practical blueprint for real-world industrial AI: \textit{generate what you cannot collect, adapt what you cannot retrain, and deploy with confidence}.

%% file: sec/7_limitations.tex
\section{Limitations}
\label{sec:limitations}

While our approach demonstrates strong performance and practical viability, several limitations stem from real-world constraints rather than conceptual shortcomings.

First, our synthetic dataset, though diverse, was capped at 2,000 images due to  resource demands of high-fidelity generation and human-in-the-loop bounding box annotation. Second, our evaluation is limited to static imagery. Many industrial hazards develop over time, and incorporating temporal cues from video could improve detection robustness. Third, while our human-in-the-loop spill placement adds essential realism, it introduces subjectivity and cannot be automated without compromising plausibility.\\

\noindent Finally, our experiments focus exclusively on vision data. In practice, multispectral inputs, such as thermal or infrared, could help disambiguate visually ambiguous cases. Supporting such modalities would require a restructured generation and adaptation pipeline.

%% file: sec/X_suppl.tex
\clearpage
\section*{Supplementary Material}

This supplementary section provides additional details, visualizations, and implementation specifics that support the main claims and methodology of our work. It is organized as follows:

\subsection*{A. Additional Qualitative Results}
\begin{figure*}[t]
    \centering
    \begin{subfigure}{0.48\textwidth}
        \centering
        \includegraphics[width=\linewidth]{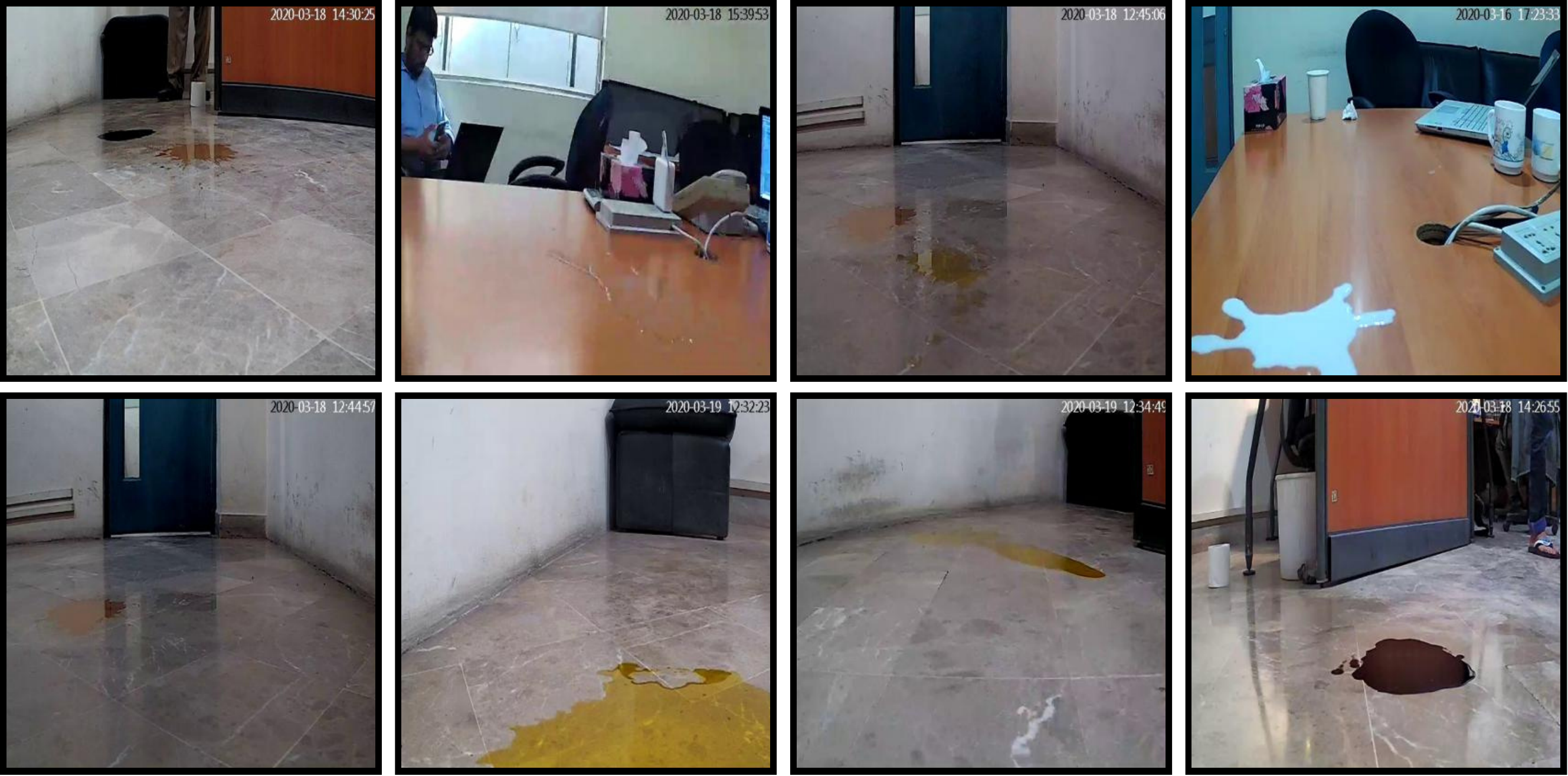}
        \caption{Public (Web Scraped) Dataset Images}
        \label{fig:pub_samples}
    \end{subfigure}
    
    \vspace{0.5em} % optional spacing between the subfigures
    \begin{subfigure}{0.48\textwidth}
    \centering
    \includegraphics[width=\linewidth]{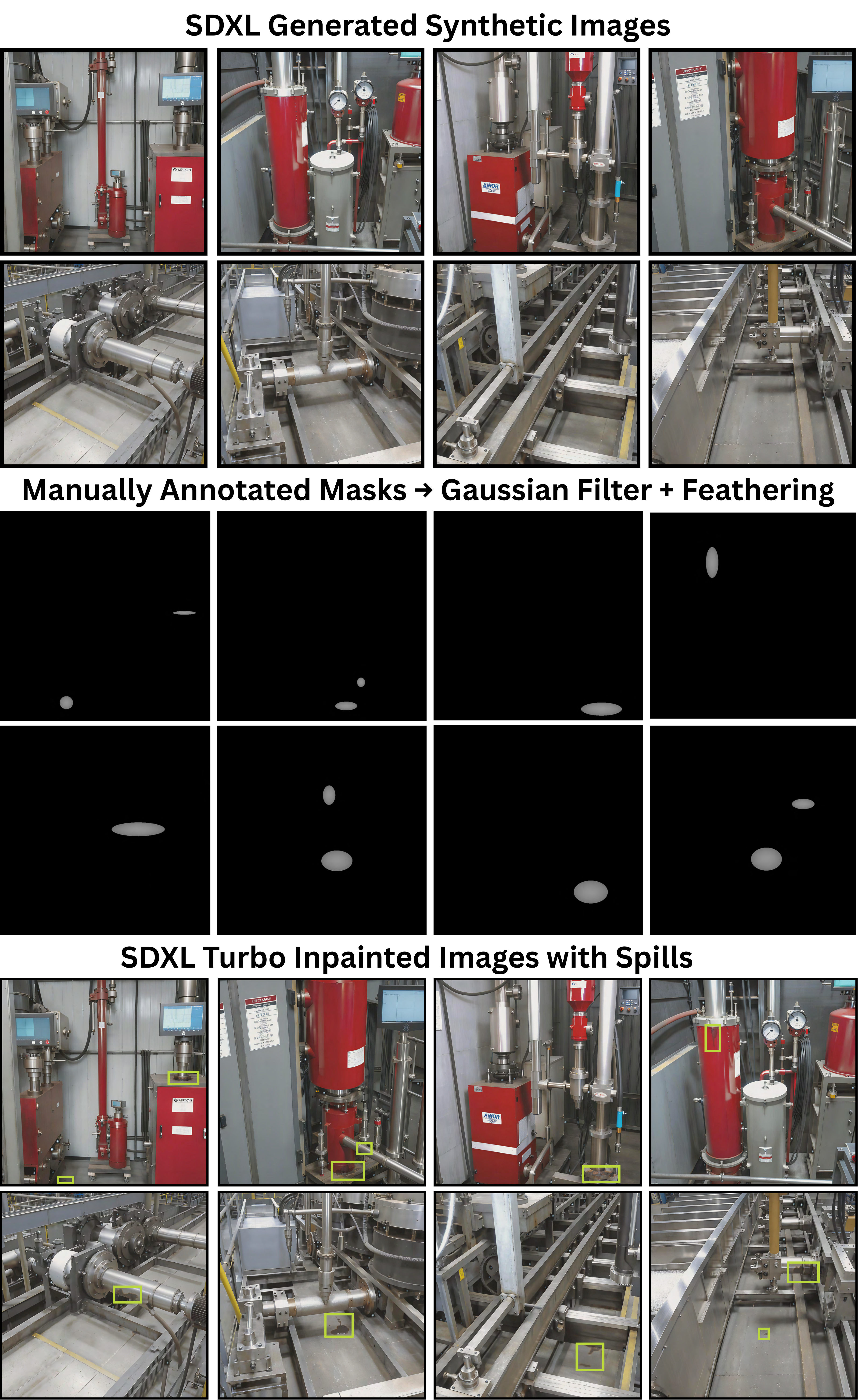}
    \caption{Synthetic Data Images}
    \label{fig:synth_samples}
    \end{subfigure}

    \caption{Dataset Samples}
    \label{fig:data_samples}
\end{figure*}

Figure\ref{fig:synth_samples} showcases additional samples from our synthetic dataset. Each row depicts the clean background, expert-annotated mask, and the final inpainted output. These examples illustrate the diversity of spill locations and materials (oil, water, rust) our pipeline can handle.

\subsection*{B. Prompt Engineering Details}
\textbf{Stable Diffusion Prompts (Scene Generation)}:
\begin{itemize}
\item Positive: \texttt{"A factory interior, close-up of industrial equipment, image captured via a colored high-quality high-end inspection camera, clear mechanical details, realistic metallic textures, authentic lighting, natural industrial setting, accurate machinery components, subtle equipment variations, realistic wear and tear."}
\item Negative: \texttt{"Text, watermark, low quality, jitter, nsfw, stickers, labels, blurred details, distorted equipment, cartoonish or unrealistic textures, unnatural colors, overly bright lighting, irrelevant objects, human presence, animals, plants, visible text, duplicated or repeated elements, unrealistic proportions, overly polished surfaces, plastic-like or artificial appearance."}
\end{itemize}

\textbf{Inpainting Prompts (Hazard Synthesis)}:
\begin{itemize}
\item Positive \texttt{"Realistic oil spill in factory with brown or black stains, industrial scene with dark oil leakage stains, brown-black oily patch on factory floor, factory oil spill with realistic black sludge, realistic factory environment with oil smears, black or brown oil leakage on industrial surface, dirty oil-stained floor in realistic factory, blackened spill area in a manufacturing plant, authentic oil spill marks on brown concrete, industrial realism with black or brown oil spill."}
\item Negative \texttt{"Cartoon, anime, illustration, painting, drawing, lowres, blurry, pixelated, overexposed, unrealistic, stylized, clipart, animated, text, watermark, signature, frame, border, extra limbs, distorted hands, shiny, plastic, toy-like, glossy, yellow tint, white overlay, newspaper texture, poster art, human figures, fingers, deformed body parts, 3D render, CGI, artifact, sketch."}
\end{itemize}

\textbf{System Prompt for VLM:}
As detailed in the main text, the system prompt emphasized verifiability, physical cues, and the avoidance of speculation. Minor adjustments to wording (e.g., stressing "clearly visible hazard") impacted precision marginally.

\subsection*{C. Adaptation and LoRA Configuration}
All LoRA experiments used the following setup:
\begin{itemize}
\item Rank: 8
\item Scaling factor $\alpha$: 1/8
\item Optimizer: AdamW, learning rate 5e-5
\item Hardware: NVIDIA A100 80GB
\end{itemize}

\textbf{Annotation Sample (COCO-style)}:
\begin{verbatim}
{
"image_id": 134,
"category_id": 3,
"bbox": [256, 411, 142, 95],
"score": 0.97
}
\end{verbatim}

\subsection*{D. Extended Evaluation Trends}
In Table\ref{tab:hit_rates_se}, we provide hit-rates at stricter IoU thresholds for Qwen-7B across adaptation strategies. These reinforce that models adapted using LoRA (especially on vision pathways) maintain more accurate localization under tighter criteria.

\begin{table}[hbt!]
\centering
\caption{Hit-rate of Qwen-7B under varying IoU thresholds}
\label{tab:iou_trends}
\begin{tabular}{lccccc}
\toprule
Method & 0.5 & 0.6 & 0.7 & 0.8 & 0.9 \\
\midrule
Zero-Shot     & 0.35 & 0.22 & 0.15 & 0.08 & 0.02 \\
ICL (10-shot) & 0.53 & 0.41 & 0.29 & 0.17 & 0.05 \\
LoRA (V)      & 0.58 & 0.49 & 0.38 & 0.26 & 0.10 \\
LoRA (V+L)    & \textbf{0.63} & \textbf{0.54} & \textbf{0.42} & \textbf{0.29} & \textbf{0.13} \\
\bottomrule
\end{tabular}
\end{table}

\subsection*{E. Dataset License and Reuse}
We release SynthSpill dataset under the \textbf{CC BY-NC 4.0 license}. Researchers may:
\begin{itemize}
\item Use the data for non-commercial research purposes
\item Modify or build upon the dataset
\item Share derived work with attribution
\end{itemize}

\subsection*{F. Visuals}

%% file: main.bbl
\begin{thebibliography}{56}
\providecommand{\natexlab}[1]{#1}
\providecommand{\url}[1]{\texttt{#1}}
\expandafter\ifx\csname urlstyle\endcsname\relax
  \providecommand{\doi}[1]{doi: #1}\else
  \providecommand{\doi}{doi: \begingroup \urlstyle{rm}\Url}\fi

\bibitem[Adams and Bloom(1995)]{adams1995pyramid}
John Adams and Howard Bloom.
\newblock Gaussian pyramid image and its application to change detection.
\newblock \emph{Computer Vision and Image Understanding}, 1995.

\bibitem[Bai et~al.(2025)Bai, Chen, Liu, Wang, Ge, Song, Dang, Wang, Wang,
  Tang, Zhong, Zhu, Yang, Li, Wan, Wang, Ding, Fu, Xu, Ye, Zhang, Xie, Cheng,
  Zhang, Yang, Xu, and Lin]{bai2025qwen25vltechnicalreport}
Shuai Bai, Keqin Chen, Xuejing Liu, Jialin Wang, Wenbin Ge, Sibo Song, Kai
  Dang, Peng Wang, Shijie Wang, Jun Tang, Humen Zhong, Yuanzhi Zhu, Mingkun
  Yang, Zhaohai Li, Jianqiang Wan, Pengfei Wang, Wei Ding, Zheren Fu, Yiheng
  Xu, Jiabo Ye, Xi Zhang, Tianbao Xie, Zesen Cheng, Hang Zhang, Zhibo Yang,
  Haiyang Xu, and Junyang Lin.
\newblock Qwen2.5-vl technical report, 2025.

\bibitem[Brown and et~al.(2020)]{brown2020language}
Tom Brown and et al.
\newblock Language models are few-shot learners.
\newblock In \emph{NeurIPS}, 2020.

\bibitem[Carion et~al.(2020)Carion, Massa, Synnaeve, Usunier, Kirillov, and
  Zagoruyko]{carion2020endtoendobjectdetectiontransformers}
Nicolas Carion, Francisco Massa, Gabriel Synnaeve, Nicolas Usunier, Alexander
  Kirillov, and Sergey Zagoruyko.
\newblock End-to-end object detection with transformers, 2020.

\bibitem[Chen et~al.(2022{\natexlab{a}})Chen, Ge, Tong, Wang, Song, Wang, and
  Luo]{chen2022adaptformeradaptingvisiontransformers}
Shoufa Chen, Chongjian Ge, Zhan Tong, Jiangliu Wang, Yibing Song, Jue Wang, and
  Ping Luo.
\newblock Adaptformer: Adapting vision transformers for scalable visual
  recognition, 2022{\natexlab{a}}.

\bibitem[Chen et~al.(2024)Chen, Lin, Zhou, Huang, Jia, Cao, and
  Wen]{chen2024iclevalevaluatingincontextlearning}
Wentong Chen, Yankai Lin, ZhenHao Zhou, HongYun Huang, Yantao Jia, Zhao Cao,
  and Ji-Rong Wen.
\newblock Icleval: Evaluating in-context learning ability of large language
  models, 2024.

\bibitem[Chen et~al.(2022{\natexlab{b}})Chen, Hu, Jin, Li, and
  Wang]{chen2022understandingdomainrandomizationsimtoreal}
Xiaoyu Chen, Jiachen Hu, Chi Jin, Lihong Li, and Liwei Wang.
\newblock Understanding domain randomization for sim-to-real transfer,
  2022{\natexlab{b}}.

\bibitem[Dettmers and et~al.(2023)]{dettmers2023qlora}
Tim Dettmers and et al.
\newblock Qlora: Efficient finetuning of quantized llms.
\newblock \emph{arXiv preprint arXiv:2310.02578}, 2023.

\bibitem[Dosovitskiy et~al.(2017)Dosovitskiy, Ros, Codevilla, Lopez, and
  Koltun]{dosovitskiy2017carlaopenurbandriving}
Alexey Dosovitskiy, German Ros, Felipe Codevilla, Antonio Lopez, and Vladlen
  Koltun.
\newblock Carla: An open urban driving simulator, 2017.

\bibitem[Hendrycks et~al.(2022)Hendrycks, Zou, Mazeika, Tang, Li, Song, and
  Steinhardt]{hendrycks2022pixmixdreamlikepicturescomprehensively}
Dan Hendrycks, Andy Zou, Mantas Mazeika, Leonard Tang, Bo Li, Dawn Song, and
  Jacob Steinhardt.
\newblock Pixmix: Dreamlike pictures comprehensively improve safety measures,
  2022.

\bibitem[Hidayatullah et~al.(2025)Hidayatullah, Syakrani, Sholahuddin, Gelar,
  and Tubagus]{hidayatullah2025yolov8yolo11comprehensivearchitecture}
Priyanto Hidayatullah, Nurjannah Syakrani, Muhammad~Rizqi Sholahuddin, Trisna
  Gelar, and Refdinal Tubagus.
\newblock Yolov8 to yolo11: A comprehensive architecture in-depth comparative
  review, 2025.

\bibitem[Ho et~al.(2020)Ho, Jain, and
  Abbeel]{ho2020denoisingdiffusionprobabilisticmodels}
Jonathan Ho, Ajay Jain, and Pieter Abbeel.
\newblock Denoising diffusion probabilistic models, 2020.

\bibitem[Hu et~al.(2022)Hu, Shen, Wallis, Allen-Zhu, Li, Wang, and
  Chen]{hu2022lora}
Edward Hu, Yelong Shen, Phillip Wallis, Zeyuan Allen-Zhu, Yuanzhi Li, Shean
  Wang, and Weizhu Chen.
\newblock Lora: Low-rank adaptation of large language models.
\newblock In \emph{ICLR}, 2022.

\bibitem[Jegham et~al.(2025)Jegham, Koh, Abdelatti, and
  Hendawi]{jegham2025yoloevolutioncomprehensivebenchmark}
Nidhal Jegham, Chan~Young Koh, Marwan Abdelatti, and Abdeltawab Hendawi.
\newblock Yolo evolution: A comprehensive benchmark and architectural review of
  yolov12, yolo11, and their previous versions, 2025.

\bibitem[Jia et~al.(2021)Jia, Yang, Xia, Chen, Parekh, Pham, Le, Sung, Li, and
  Duerig]{jia2021scalingvisualvisionlanguagerepresentation}
Chao Jia, Yinfei Yang, Ye Xia, Yi-Ting Chen, Zarana Parekh, Hieu Pham, Quoc~V.
  Le, Yunhsuan Sung, Zhen Li, and Tom Duerig.
\newblock Scaling up visual and vision-language representation learning with
  noisy text supervision, 2021.

\bibitem[Jia et~al.(2022)Jia, Tang, Chen, Cardie, Belongie, Hariharan, and
  Lim]{jia2022vpt}
Menglin Jia, Luming Tang, Bor-Chun Chen, Claire Cardie, Serge Belongie, Bharath
  Hariharan, and Ser-Nam Lim.
\newblock Visual prompt tuning, 2022.

\bibitem[Jocher et~al.(2023)Jocher, Chaurasia, and Qiu]{yolov8}
Glenn Jocher, Ayush Chaurasia, and Jing Qiu.
\newblock Ultralytics yolov8.
\newblock \url{https://github.com/ultralytics/ultralytics}, 2023.

\bibitem[Karras et~al.(2020)Karras, Laine, Aittala, Hellsten, Lehtinen, and
  Aila]{karras2020analyzingimprovingimagequality}
Tero Karras, Samuli Laine, Miika Aittala, Janne Hellsten, Jaakko Lehtinen, and
  Timo Aila.
\newblock Analyzing and improving the image quality of stylegan, 2020.

\bibitem[Kirillov et~al.(2023{\natexlab{a}})Kirillov, Mintun, Ravi, Mao,
  Rolland, Gustafson, Xiao, Whitehead, Berg, Lo, et~al.]{kirillov2023segment}
Alexander Kirillov, Eric Mintun, Nikhila Ravi, Hanzi Mao, Chloe Rolland, Laura
  Gustafson, Tete Xiao, Spencer Whitehead, Alexander~C Berg, Wan-Yen Lo, et~al.
\newblock Segment anything.
\newblock In \emph{Proceedings of the IEEE/CVF international conference on
  computer vision}, pages 4015--4026, 2023{\natexlab{a}}.

\bibitem[Kirillov et~al.(2023{\natexlab{b}})Kirillov, Mintun, Ravi, Mao,
  Rolland, Gustafson, Romero, Krainin, Li, Li, et~al.]{kirillov2023segany}
Alexander Kirillov, Eric Mintun, Nikhila Ravi, Hanzi Mao, Lily Rolland, Laura
  Gustafson, Callen Romero, Michael Krainin, David Li, Chao Li, et~al.
\newblock Segment anything.
\newblock In \emph{Proceedings of the IEEE/CVF Conference on Computer Vision
  and Pattern Recognition}, pages 2426--2437, 2023{\natexlab{b}}.

\bibitem[Kirkpatrick et~al.(2017)Kirkpatrick, Pascanu, Rabinowitz, Veness,
  Desjardins, Rusu, Milan, Quan, Ramalho, Grabska-Barwinska, Hassabis, Clopath,
  Kumaran, and Hadsell]{Kirkpatrick_2017}
James Kirkpatrick, Razvan Pascanu, Neil Rabinowitz, Joel Veness, Guillaume
  Desjardins, Andrei~A. Rusu, Kieran Milan, John Quan, Tiago Ramalho, Agnieszka
  Grabska-Barwinska, Demis Hassabis, Claudia Clopath, Dharshan Kumaran, and
  Raia Hadsell.
\newblock Overcoming catastrophic forgetting in neural networks.
\newblock \emph{Proceedings of the National Academy of Sciences}, 114\penalty0
  (13):\penalty0 3521–3526, 2017.

\bibitem[Kolve et~al.(2022)Kolve, Mottaghi, Han, VanderBilt, Weihs, Herrasti,
  Deitke, Ehsani, Gordon, Zhu, Kembhavi, Gupta, and
  Farhadi]{kolve2022ai2thorinteractive3denvironment}
Eric Kolve, Roozbeh Mottaghi, Winson Han, Eli VanderBilt, Luca Weihs, Alvaro
  Herrasti, Matt Deitke, Kiana Ehsani, Daniel Gordon, Yuke Zhu, Aniruddha
  Kembhavi, Abhinav Gupta, and Ali Farhadi.
\newblock Ai2-thor: An interactive 3d environment for visual ai, 2022.

\bibitem[{laion}(2023)]{laion_clip_vith14_laion2b_2023}
{laion}.
\newblock {CLIP ViT-H/14 – LAION‑2B
  (laion/CLIP‑ViT‑H‑14‑laion2B‑s32B‑b79K)}.
\newblock \url{https://huggingface.co/laion/CLIP-ViT-H-14-laion2B-s32B-b79K},
  2023.
\newblock Published ca. Sep 2023; Accessed: 2025‑07‑04.

\bibitem[Levin and Fried(2024)]{levin2024differentialdiffusiongivingpixel}
Eran Levin and Ohad Fried.
\newblock Differential diffusion: Giving each pixel its strength, 2024.

\bibitem[Li et~al.(2022)Li, Zhang, Zhang, Yang, Li, Zhong, Wang, Yuan, Zhang,
  Hwang, et~al.]{li2022grounded}
Liunian~Harold Li, Pengchuan Zhang, Haotian Zhang, Jianwei Yang, Chunyuan Li,
  Yiwu Zhong, Lijuan Wang, Lu Yuan, Lei Zhang, Jenq-Neng Hwang, et~al.
\newblock Grounded language-image pre-training.
\newblock In \emph{Proceedings of the IEEE/CVF Conference on Computer Vision
  and Pattern Recognition}, pages 10965--10975, 2022.

\bibitem[Lian et~al.(2023)Lian, Zhou, Feng, and
  Wang]{lian2023scalingshiftingfeatures}
Dongze Lian, Daquan Zhou, Jiashi Feng, and Xinchao Wang.
\newblock Scaling \& shifting your features: A new baseline for efficient model
  tuning, 2023.

\bibitem[Lin et~al.(2025)Lin, Li, Yao, Dong, Guo, Hong, Zhang, and
  Wen]{lin2025generalizationenhancedfewshotobjectdetection}
Hui Lin, Nan Li, Pengjuan Yao, Kexin Dong, Yuhan Guo, Danfeng Hong, Ying Zhang,
  and Congcong Wen.
\newblock Generalization-enhanced few-shot object detection in remote sensing,
  2025.

\bibitem[Liu et~al.(2024)Liu, Wei, Liu, Si, Zhang, Rao, Zheng, Peng, Yang,
  Zhou, and Dai]{liu2024bestpracticeslessonslearned}
Ruibo Liu, Jerry Wei, Fangyu Liu, Chenglei Si, Yanzhe Zhang, Jinmeng Rao,
  Steven Zheng, Daiyi Peng, Diyi Yang, Denny Zhou, and Andrew~M. Dai.
\newblock Best practices and lessons learned on synthetic data, 2024.

\bibitem[Liu and et~al.(2023)]{liu2023grounding}
Xin Liu and et al.
\newblock Grounding dino: Marrying object detection with grounded language
  queries.
\newblock In \emph{CVPR}, 2023.

\bibitem[Liu et~al.(2021)Liu, Lin, Cao, Hu, Wei, Zhang, Lin, and
  Guo]{liu2021swintransformerhierarchicalvision}
Ze Liu, Yutong Lin, Yue Cao, Han Hu, Yixuan Wei, Zheng Zhang, Stephen Lin, and
  Baining Guo.
\newblock Swin transformer: Hierarchical vision transformer using shifted
  windows, 2021.

\bibitem[Lugmayr et~al.(2022)Lugmayr, Danelljan, Romero, Yu, Timofte, and
  Gool]{lugmayr2022repaintinpaintingusingdenoising}
Andreas Lugmayr, Martin Danelljan, Andres Romero, Fisher Yu, Radu Timofte, and
  Luc~Van Gool.
\newblock Repaint: Inpainting using denoising diffusion probabilistic models,
  2022.

\bibitem[Luo et~al.(2023)Luo, Tan, Patil, Gu, von Platen, Passos, Huang, Li,
  and Zhao]{luo2023lcmlorauniversalstablediffusionacceleration}
Simian Luo, Yiqin Tan, Suraj Patil, Daniel Gu, Patrick von Platen, Apolinário
  Passos, Longbo Huang, Jian Li, and Hang Zhao.
\newblock Lcm-lora: A universal stable-diffusion acceleration module, 2023.

\bibitem[Mishra et~al.(2022)Mishra, Panda, Phoo, Chen, Karlinsky, Saenko,
  Saligrama, and Feris]{mishra2022task2simeffectivepretraining}
Samarth Mishra, Rameswar Panda, Cheng~Perng Phoo, Chun-Fu Chen, Leonid
  Karlinsky, Kate Saenko, Venkatesh Saligrama, and Rogerio~S. Feris.
\newblock Task2sim : Towards effective pre-training and transfer from synthetic
  data, 2022.

\bibitem[Moenck et~al.(2024)Moenck, Thieu, Koch, and
  Schüppstuhl]{moenck2024industriallanguageimagedatasetilid}
Keno Moenck, Duc~Trung Thieu, Julian Koch, and Thorsten Schüppstuhl.
\newblock Industrial language-image dataset (ilid): Adapting vision foundation
  models for industrial settings, 2024.

\bibitem[{ostris}(2024)]{ostris_ipcompositionadapter_2024}
{ostris}.
\newblock {IP Composition Adapter} (stable diffusion) model.
\newblock \url{https://huggingface.co/ostris/ip-composition-adapter}, 2024.
\newblock Published: March 20, 2024; Accessed: 2025-07-04.

\bibitem[Pfeiffer et~al.(2021)Pfeiffer, Kamath, Rücklé, Cho, and
  Gurevych]{pfeiffer2021adapterfusionnondestructivetaskcomposition}
Jonas Pfeiffer, Aishwarya Kamath, Andreas Rücklé, Kyunghyun Cho, and Iryna
  Gurevych.
\newblock Adapterfusion: Non-destructive task composition for transfer
  learning, 2021.

\bibitem[Poole et~al.(2022)Poole, Jain, Barron, and
  Mildenhall]{poole2022dreamfusiontextto3dusing2d}
Ben Poole, Ajay Jain, Jonathan~T. Barron, and Ben Mildenhall.
\newblock Dreamfusion: Text-to-3d using 2d diffusion, 2022.

\bibitem[Radford et~al.(2021)Radford, Kim, Hallacy, Ramesh, Goh, Agarwal,
  Sastry, Askell, Mishkin, Clark, Krueger, and
  Sutskever]{radford2021learningtransferablevisualmodels}
Alec Radford, Jong~Wook Kim, Chris Hallacy, Aditya Ramesh, Gabriel Goh,
  Sandhini Agarwal, Girish Sastry, Amanda Askell, Pamela Mishkin, Jack Clark,
  Gretchen Krueger, and Ilya Sutskever.
\newblock Learning transferable visual models from natural language
  supervision, 2021.

\bibitem[Radke et~al.(2005)Radke, Andra, Al-Kofahi, and Roysam]{radke2005image}
R.~J. Radke, S. Andra, O. Al-Kofahi, and B. Roysam.
\newblock Image change detection algorithms: A systematic survey.
\newblock \emph{IEEE Transactions on Image Processing}, 14\penalty0
  (3):\penalty0 294--307, 2005.

\bibitem[Ramesh et~al.(2022)Ramesh, Dhariwal, Nichol, Chu, and
  Chen]{ramesh2022hierarchicaltextconditionalimagegeneration}
Aditya Ramesh, Prafulla Dhariwal, Alex Nichol, Casey Chu, and Mark Chen.
\newblock Hierarchical text-conditional image generation with clip latents,
  2022.

\bibitem[Redmon and Farhadi(2018)]{redmon2018yolov3}
Joseph Redmon and Ali Farhadi.
\newblock Yolov3: An incremental improvement.
\newblock \emph{arXiv preprint arXiv:1804.02767}, 2018.

\bibitem[Ren et~al.(2015)Ren, He, Girshick, and Sun]{ren2015faster}
Shaoqing Ren, Kaiming He, Ross Girshick, and Jian Sun.
\newblock Faster r-cnn: Towards real-time object detection with region proposal
  networks.
\newblock In \emph{NeurIPS}, 2015.

\bibitem[Rombach et~al.(2022)Rombach, Blattmann, Lorenz, Esser, and
  Ommer]{rombach2022highresolutionimagesynthesislatent}
Robin Rombach, Andreas Blattmann, Dominik Lorenz, Patrick Esser, and Björn
  Ommer.
\newblock High-resolution image synthesis with latent diffusion models, 2022.

\bibitem[Ruiz et~al.(2023)Ruiz, Li, Jampani, Pritch, Rubinstein, and
  Aberman]{ruiz2023dreamboothfinetuningtexttoimage}
Nataniel Ruiz, Yuanzhen Li, Varun Jampani, Yael Pritch, Michael Rubinstein, and
  Kfir Aberman.
\newblock Dreambooth: Fine tuning text-to-image diffusion models for
  subject-driven generation, 2023.

\bibitem[Saleh et~al.(2021)Saleh, Szenasi, and Vamossy]{Saleh_2021}
Kaziwa Saleh, Sandor Szenasi, and Zoltan Vamossy.
\newblock Occlusion handling in generic object detection: A review.
\newblock In \emph{2021 IEEE 19th World Symposium on Applied Machine
  Intelligence and Informatics (SAMI)}, page 000477–000484. IEEE, 2021.

\bibitem[Sapkota et~al.(2025)Sapkota, Cheppally, Sharda, and
  Karkee]{sapkota2025rfdetrobjectdetectionvs}
Ranjan Sapkota, Rahul~Harsha Cheppally, Ajay Sharda, and Manoj Karkee.
\newblock Rf-detr object detection vs yolov12 : A study of transformer-based
  and cnn-based architectures for single-class and multi-class greenfruit
  detection in complex orchard environments under label ambiguity, 2025.

\bibitem[{Stability AI}(2023)]{stabilityai_sdxl-turbo_fp16_2023}
{Stability AI}.
\newblock {SDXL‑Turbo 1.0 fp16} model checkpoint.
\newblock
  \url{https://huggingface.co/stabilityai/sdxl-turbo/blob/main/sd_xl_turbo_1.0_fp16.safetensors},
  2023.
\newblock Accessed: 2025-07-04.

\bibitem[{Unknown Author}(2025)]{interiorSceneXL2025}
{Unknown Author}.
\newblock Interior scene xl (stable diffusion xl model).
\newblock \url{https://civitai.com/models/715747/interior-scene-xl}, 2025.
\newblock Accessed: 2025-07-04.

\bibitem[Wang et~al.(2022)Wang, Bochkovskiy, and
  Liao]{wang2022yolov7trainablebagoffreebiessets}
Chien-Yao Wang, Alexey Bochkovskiy, and Hong-Yuan~Mark Liao.
\newblock Yolov7: Trainable bag-of-freebies sets new state-of-the-art for
  real-time object detectors, 2022.

\bibitem[Wang et~al.(2021)Wang, Liu, Yan, and Tang]{wang2021survey}
Ping Wang, Jiangbo Liu, Yilai Yan, and Zongjian Tang.
\newblock A survey on deep learning-based industrial defect detection.
\newblock \emph{IEEE Transactions on Neural Networks and Learning Systems},
  2021.

\bibitem[Ye et~al.(2023)Ye, Zhang, Liu, Han, and
  Yang]{ye2023ipadaptertextcompatibleimage}
Hu Ye, Jun Zhang, Sibo Liu, Xiao Han, and Wei Yang.
\newblock Ip-adapter: Text compatible image prompt adapter for text-to-image
  diffusion models, 2023.

\bibitem[Yuan et~al.(2021)Yuan, Chen, Chen, Codella, Dai, Gao, Hu, Huang, Li,
  Li, Liu, Liu, Liu, Lu, Shi, Wang, Wang, Xiao, Xiao, Yang, Zeng, Zhou, and
  Zhang]{yuan2021florencenewfoundationmodel}
Lu Yuan, Dongdong Chen, Yi-Ling Chen, Noel Codella, Xiyang Dai, Jianfeng Gao,
  Houdong Hu, Xuedong Huang, Boxin Li, Chunyuan Li, Ce Liu, Mengchen Liu,
  Zicheng Liu, Yumao Lu, Yu Shi, Lijuan Wang, Jianfeng Wang, Bin Xiao, Zhen
  Xiao, Jianwei Yang, Michael Zeng, Luowei Zhou, and Pengchuan Zhang.
\newblock Florence: A new foundation model for computer vision, 2021.

\bibitem[Zaken et~al.(2022)Zaken, Ravfogel, and
  Goldberg]{zaken2022bitfitsimpleparameterefficientfinetuning}
Elad~Ben Zaken, Shauli Ravfogel, and Yoav Goldberg.
\newblock Bitfit: Simple parameter-efficient fine-tuning for transformer-based
  masked language-models, 2022.

\bibitem[Zhang et~al.(2022)Zhang, Li, Liu, Zhang, Su, Zhu, Ni, and
  Shum]{zhang2022dinodetrimproveddenoising}
Hao Zhang, Feng Li, Shilong Liu, Lei Zhang, Hang Su, Jun Zhu, Lionel~M. Ni, and
  Heung-Yeung Shum.
\newblock Dino: Detr with improved denoising anchor boxes for end-to-end object
  detection, 2022.

\bibitem[Zhang et~al.(2023)Zhang, Rao, and
  Agrawala]{zhang2023addingconditionalcontroltexttoimage}
Lvmin Zhang, Anyi Rao, and Maneesh Agrawala.
\newblock Adding conditional control to text-to-image diffusion models, 2023.

\bibitem[Zhou et~al.(2024)Zhou, Li, Wang, and
  Shen]{zhou2024visualincontextlearninglarge}
Yucheng Zhou, Xiang Li, Qianning Wang, and Jianbing Shen.
\newblock Visual in-context learning for large vision-language models, 2024.

\end{thebibliography}
